\documentclass[11pt]{article}
\usepackage{acl}
\usepackage{times,latexsym,booktabs,graphicx}
\usepackage[T1]{fontenc}
\usepackage[utf8]{inputenc}
\usepackage{kotex}
\usepackage{longtable,array,fvextra}
\input{omlcmm.fd}
\DeclareFontShape{OML}{cmm}{m}{it}{<5><6><7><8>gen*cmmi<9><10><10.95>cmmi10<12><14.4><17.28><20.74><24.88>cmmi12}{}

\title{Same Rule, Different Verdicts: Rubric Language and Clause Structure\\
in LLM Judges for Korean Support Dialogues}
\author{Sangdo Han \\
  CEO Business School \\
  \texttt{sangdohan@ceobizschool.kr} \\\And
  Moonsoo Kim\thanks{Corresponding author.} \\
  CEO Business School \\
  \texttt{mskim@ceobizschool.kr}}

\begin{document}
\maketitle

\begin{abstract}
Services that operate in Korean often write their evaluation rubrics in Korean
and hand them to an LLM judge. We test whether a judge's verdict depends on the
language of the rubric and on how a rubric item is phrased. We write 12
criteria for support dialogues in four clause structures (affirmative, negated,
conditional, exception) such that two pairs of structures have the same
meaning, and we give each item in Korean and in English. Six low-cost models
from three providers, plus one reference model, judge 60 translated ESConv
dialogues and 144 constructed boundary cases under every combination
(51,744 requests across the seven models). The
analysis plan was fixed and committed before data collection. Cross-provider
agreement is lower with Korean rubrics than with English rubrics (Cohen's
$\kappa$ 0.687 vs.\ 0.745; difference $-0.058$, 95\% CI $[-0.072,-0.045]$).
Disagreement concentrates in negated, conditional, and exception items (odds
ratio 5.6 relative to affirmative items). A single model changes its verdict
between two items with the same meaning in 11--15\% of cases. Two hypotheses
are not supported: models from the same provider do not agree more than models
from different providers, and adding paired examples does not reduce the
language gap. We measure consistency only; no human labels are used, and all
materials were drafted by an LLM.
\end{abstract}

\section{Introduction}

\citet{zheng2023judging} found that strong LLMs judging open-ended chat
responses agree with human preferences about as well as humans agree with each
other, which makes an LLM judge an attractive substitute for human evaluation of
dialogue systems. In
non-English deployments the rubric is usually written in the service language,
because the people who own the rubric read and maintain it in that language.
Whether this choice changes the verdicts is an empirical question with a
practical cost: if two judges read the same Korean item differently, a rubric
score reflects the choice of judge as much as the dialogue.

Prior work has shown that multilingual judges are inconsistent across languages
\citep{fu2025reliable}, that English instructions can give more consistent
judgments than translated ones \citep{zubiaga2026reliable}, and that the
language of a judge prompt interacts with the judge model
\citep{mahmood2026localization}. These studies vary the language. They do not
control how an item is phrased. Rubrics, however, are full of negations,
conditions, and exceptions, and these are the constructions in which a small
difference in reading changes the verdict.

We control both factors. Each criterion is written in four structures, and the
structures form two pairs with the same meaning. This gives a test that needs
no reference labels: if one model gives different verdicts to two items that
mean the same thing, at least one verdict is wrong. We ask four
questions, fixed before data collection (Section~\ref{sec:plan}):

\begin{itemize}
\item \textbf{H1.} Cross-provider agreement is lower with Korean rubrics than
with English rubrics.
\item \textbf{H2.} Disagreement is concentrated in negated, conditional, and
exception items.
\item \textbf{H3.} Agreement within a provider is higher than agreement across
providers.
\item \textbf{H4.} Paired examples reduce the language gap.
\end{itemize}

H1 and H2 are supported. H3 and H4 are not. We report all four.

\section{Related Work}

\paragraph{Multilingual LLM judges.}
\citet{fu2025reliable} measure the consistency of judge verdicts across 25
languages on parallel data and report low agreement. \citet{zubiaga2026reliable}
compare translated and English rubrics for fine-tuned judges in English,
Spanish, and Basque. \citet{mahmood2026localization} localize a judge prompt
into five languages for requirement-level judgments of code agents and report
that judge rankings change with language and that inter-judge agreement is low
in every language. Related efforts build multilingual meta-evaluation benchmarks for judges
\citep{son2025mmevalmultilingualmetaevaluationbenchmark}, train evaluators that
score non-English responses against English references
\citep{doddapaneni-etal-2025-cross}, and transfer judges across languages by
decomposing evaluation into shared criteria
\citep{sheth2026crosslingualllmjudgetransferevaluation}. Our H1 and H3 extend
these observations to Korean and to support dialogues. To our knowledge, none of these studies controls the clause structure of the
rubric items or uses meaning-equivalent item pairs.

\paragraph{Rubric wording.}
\citet{rubricartifacts2026} show that a classifier given only the text of an
English rubric item, without the response, predicts a judge's verdict well above
chance, and that negation words in an item are associated with negative
verdicts. They also reverse the meaning of rubric criteria and find that the
judge often keeps its verdict. That design changes the meaning and asks whether
the verdict follows. Ours keeps the meaning fixed,
changes only the structure, and asks whether the verdict stays. The two designs
are complementary.

\paragraph{Negation and prompt wording.}
Language models perform poorly on negation benchmarks
\citep{truong2023naysayers}, and in few-shot settings their accuracy changes
substantially with formatting choices that leave the meaning of a prompt
unchanged \citep{sclar2024quantifying}.
Rubric items with negations, conditions, and exceptions combine both risks.
Our structure contrasts measure them in the setting where a judge must apply
such an item to a dialogue.

\section{Method}

\subsection{Rubric}

We define 12 criteria for the helper's behaviour in a support dialogue (for
example, asking about the client's situation before giving the first piece of
advice). Each criterion is written as four items:

\begin{itemize}
\item \textbf{POS}: an affirmative requirement.
\item \textbf{NEG}: a negated sentence with the same satisfaction conditions as
POS.
\item \textbf{COND}: the requirement under a condition (``if the client has not
asked for advice, \dots'').
\item \textbf{EXC}: the requirement followed by an exception (``\dots This does
not apply if the client has asked for advice.'').
\end{itemize}

POS and NEG mean the same thing; COND and EXC mean the same thing. Each item
exists in Korean and in English, giving 96 items. The judge instructions are
also written in both languages and are in the language of the rubric. The
instructions define the three verdicts (met, violated, not applicable). For the
example condition, each item comes with one satisfying and one violating
example; the four structures of a criterion share the same example text.

All items, examples, and instructions were drafted by Claude and were not
reviewed by a human before data collection. Equivalence between the Korean and
English items was checked by back-translation with two models from different
providers (Claude Sonnet 5.5 and GPT-5.4 nano); the comparison itself was also
done by Claude. Items were revised twice on this basis and all 96 were kept.
Three items were mistranslated by one back-translation model only and are
listed in the appendix.

\subsection{Data}

\paragraph{Set A: natural dialogues.} We sample 60 dialogues (1,721 utterances,
21--40 utterances each, 13 topics) from ESConv \citep{liu2021esconv} with a
fixed seed and translate them into Korean with Claude Sonnet 5.5. An LLM
spot-check of 12 dialogues (376 utterances) marked 5 utterances as errors
(1.3\%). No human checked the translations.

\paragraph{Set B: boundary cases.} For each criterion we construct 12 short
dialogues in Korean with English counterparts (144 cases). Cases are minimal
pairs: within a criterion the helper's turns are fixed and one client sentence
changes. The four categories are: no exception situation; the client explicitly
denies the exception situation; the client explicitly states it; the client
states it indirectly. For the last category (36 cases) a literal reading and
the reading intended by the item writer give different verdicts on COND and EXC
items. These labels come from the item writer and are not human gold labels.

\subsection{Judges and Procedure}

The confirmatory judges are six low-cost models, two from each of three
providers: Claude Sonnet 5.5 and Claude Haiku 4.5; GPT-5.6 Luna and GPT-5.4
nano; Gemini 3.5 Flash-Lite and Gemini 3.1 Flash-Lite. Claude Opus 5.5 is run
as a reference model and is excluded from all tests. Reasoning is disabled or
set to the lowest available level (Appendix~C); it cannot be disabled for the
reference model.
Default temperature is used. The same prompt text is sent to every
model, with no provider-specific structured-output feature.

A Set A request contains one Korean dialogue and the 12 items of one structure
in one rubric language. A Set B request contains one case and one item. Set B
is run with and without examples, with Korean and English cases, with a
verdict-only baseline (no rationale), and with two extra repetitions on 36
cases. Each model receives 7,392 requests; the seven models receive 51,744.
Malformed outputs are retried once and then counted as missing (765 retries
in total). Data were
collected on 6--7 October 2026 through the providers' batch interfaces.

Of 88,704 item-level verdicts, 1,219 (1.4\%) are missing. 1,199 of these are
from the reference model, almost all because the response reached the output
token limit; its reasoning cannot be disabled and counts toward that limit
(Appendix~E). The six confirmatory models have 20 missing verdicts out of
76,032 (0.03\%).

\subsection{Measures and Analysis Plan}
\label{sec:plan}

Agreement between two models is Cohen's $\kappa$ \citep{cohen1960} over the
three verdicts. There
are 15 pairs of confirmatory models: 3 within a provider and 12 across
providers. A \emph{structure flip} is a case in which one model gives different
verdicts to the POS and NEG items, or to the COND and EXC items, for the same
unit under otherwise identical conditions. A \emph{language flip} is the same
for the Korean and English versions of one item.

H1 is the difference in mean cross-provider $\kappa$ between Korean and English
rubrics on Korean units without examples (Sets A and B), with a cluster
bootstrap over units stratified by set (10,000 resamples). H2 and H4 are
contrasts in a mixed-effects logistic regression of pairwise disagreement on
rubric language, structure, pair type, and examples, with random intercepts for
unit, criterion, and model pair. H3 is the bootstrap difference between mean
within-provider and mean cross-provider $\kappa$. H2--H4 are Holm-corrected \citep{holm1979}. The regression is fitted with
lme4 \citep{lme4} and the contrasts with emmeans \citep{emmeans}; the full
model converged without simplification.

The hypotheses, estimators, tests, exclusion rules, and model settings were
fixed in a document committed and pushed to a public repository 12 minutes
before the first request was submitted. The plan was not registered with a
registry. One operational rule was added after collection began: a request that
failed with a transient provider error was resubmitted once (one request). A
planning simulation gave a minimum detectable $\kappa$ difference of
0.03--0.11 at 80\% power, depending on the assumed within-unit error
correlation.

\section{Results}

\subsection{Rubric Language (H1)}

Mean cross-provider $\kappa$ is 0.687 (95\% CI $[0.667,0.705]$) with Korean
rubrics and 0.745 ($[0.726,0.763]$) with English rubrics. The difference is
$-0.058$ ($[-0.072,-0.045]$; none of 10,000 bootstrap replicates is
non-negative). H1 is supported. The difference has the same sign in Set A
($-0.046$, $[-0.061,-0.031]$), in Set B ($-0.132$, $[-0.163,-0.103]$), and when
``not applicable'' is merged with ``met'' ($-0.057$, $[-0.073,-0.042]$).

\subsection{Clause Structure (H2)}

\begin{figure*}[t]
\centering
\includegraphics[width=\textwidth]{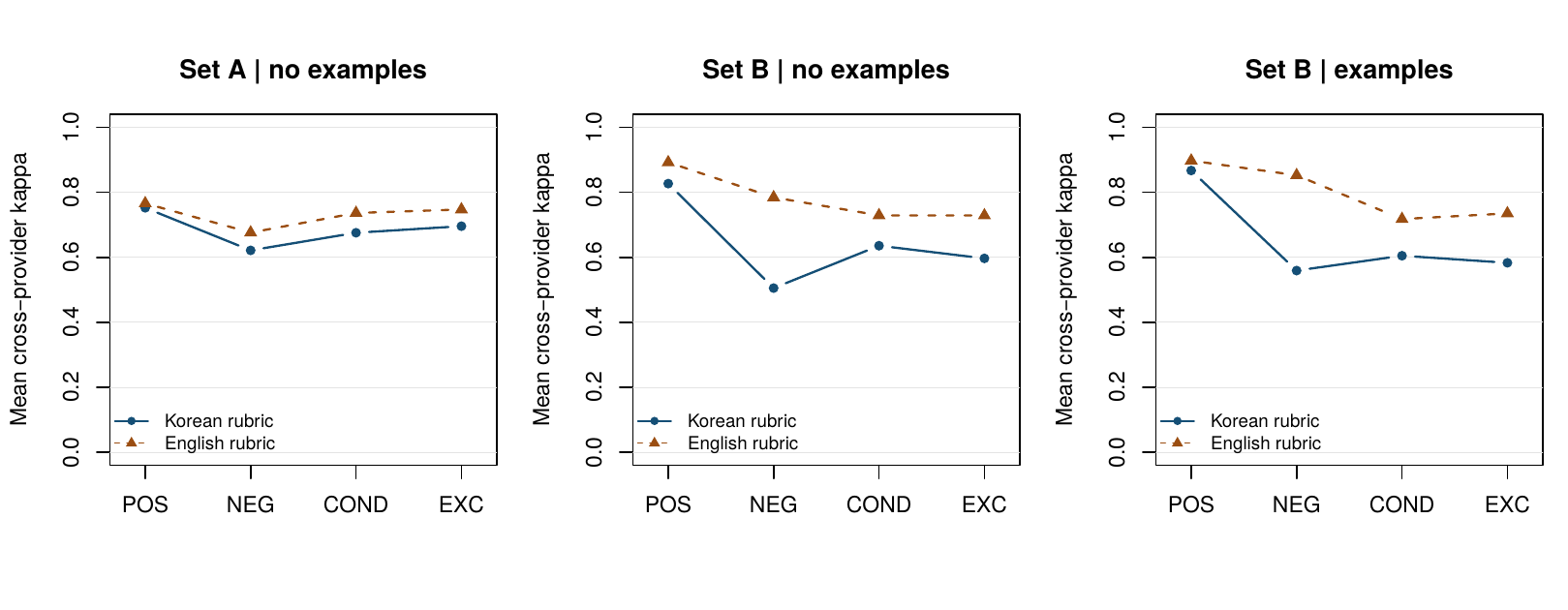}
\caption{Mean cross-provider Cohen's $\kappa$ by clause structure and rubric
language (six confirmatory models, Korean units). The two example conditions of
Set B are shown in separate panels.}
\label{fig:kappa}
\end{figure*}

The odds of disagreement for negated, conditional, and exception items,
averaged, are higher than for affirmative items (log-odds contrast 1.72,
SE 0.08, $z=21.8$, Holm-adjusted $p<10^{-100}$; odds ratio 5.6). H2 is
supported. Table~\ref{tab:structure} and Figure~\ref{fig:kappa} show the
descriptive pattern. In Set B
the largest gap between languages is for negated items: 0.51 with Korean
rubrics against 0.78 with English rubrics.

\begin{table}[t]
\centering\small
\begin{tabular}{llcccc}
\toprule
Set & Rubric & POS & NEG & COND & EXC \\
\midrule
A & Korean  & .752 & .621 & .675 & .696 \\
A & English & .766 & .676 & .736 & .747 \\
B & Korean  & .827 & .505 & .636 & .597 \\
B & English & .892 & .784 & .729 & .729 \\
\bottomrule
\end{tabular}
\caption{Mean cross-provider Cohen's $\kappa$ by structure (Korean units, no
examples, 12 model pairs).}
\label{tab:structure}
\end{table}

\subsection{Provider (H3)}

Mean within-provider $\kappa$ exceeds mean cross-provider $\kappa$ by 0.005
($[-0.001,0.012]$, Holm-adjusted $p=0.23$). H3 is not supported. Two models from
the same provider agree about as much as two models from different providers
(Korean rubrics: 0.697 within, 0.687 across; English rubrics: 0.745 and 0.745).
In Set B alone the difference is 0.021 ($[0.004,0.039]$); this is a sensitivity
analysis and does not change the conclusion.

\subsection{Paired Examples (H4)}

The language gap in disagreement does not shrink when examples are added
(log-odds contrast $-0.08$, SE 0.09, Holm-adjusted $p=0.39$). H4 is not
supported. Examples raise agreement on negated items in both languages (Set B:
Korean 0.51 to 0.56, English 0.78 to 0.85) and leave the gap in place.

\subsection{Flips Within a Model}

Across the six models, Korean units, and no examples, a model gives different
verdicts to POS and NEG items in 14.1\% of cases with Korean rubrics and 12.1\%
with English rubrics; for COND and EXC items the rates are 14.9\% and
11.0\%. 
Models differ widely. In Set B the structure-flip rate with Korean rubrics
ranges from 3.8\% (Claude Sonnet 5.5) to 24.3\% (GPT-5.4 nano).
Language flips occur in 10.8\% of cases for affirmative items and in
15.0--15.7\% for the other three structures. Table~\ref{tab:models} gives the
per-model values.

\begin{table*}[t]
\centering\small
\begin{tabular}{lcccccc}
\toprule
 & \multicolumn{2}{c}{Structure flip (\%)} & Language & Self-consist. & \multicolumn{2}{c}{Literal reading (\%)} \\
Model & Korean & English & flip (\%) & (Fleiss $\kappa$) & Korean & English \\
\midrule
Claude Sonnet 5.5     &  3.8 &  5.6 &  7.9 & .971 & 78 & 65 \\
Claude Haiku 4.5      & 15.6 & 10.1 & 15.2 & .896 & 60 & 58 \\
GPT-5.6 Luna          & 14.6 &  7.3 & 12.7 & .855 & 63 & 62 \\
GPT-5.4 nano          & 24.3 &  9.7 & 20.9 & .738 & 88 & 92 \\
Gemini 3.5 Flash-Lite & 11.5 &  5.2 & 15.4 & .937 & 43 & 47 \\
Gemini 3.1 Flash-Lite &  8.3 &  2.8 & 13.3 & .951 & 50 & 52 \\
\bottomrule
\end{tabular}
\caption{Per-model consistency. Structure flip: share of meaning-equivalent item
pairs with different verdicts (Set B, Korean and English rubrics). Language flip:
share of items with different verdicts under the Korean and English rubric.
Literal reading: share of verdicts matching the literal reading on the 36
indirectly stated cases.}
\label{tab:models}
\end{table*}

\subsection{Literal and Intended Readings}

On the 36 cases where the client states the exception situation indirectly,
the share of verdicts matching the literal reading ranges from 43\% to 92\%
across models. GPT-5.4 nano follows the literal reading in about 90\% of
verdicts in both languages; the two Gemini models are near 50\%. Claude Sonnet
5.5 follows the literal reading more often with Korean rubrics (78\%) than with
English rubrics (65\%).

\subsection{Self-Consistency and Baselines}

Fleiss' $\kappa$ over three repetitions of the same request ranges from 0.74
(GPT-5.4 nano) to 0.97 (Claude Sonnet 5.5); five of six models are at 0.85 or
above. Run-to-run variation therefore accounts for part of the cross-model
disagreement but not most of it.

Without a rationale, mean cross-provider $\kappa$ on the matched conditions is
0.55 (Korean) and 0.71 (English); with a rationale written before the verdict
it is 0.64 and 0.78. 
Asking for a rationale raises agreement in both languages and does not close
the gap. With English cases instead of Korean cases the language difference is
larger ($-0.198$, $[-0.238,-0.159]$).

\subsection{Exploratory Analyses}

The analyses in this subsection were not part of the fixed plan and involve no
significance tests.

\paragraph{Criteria.} The drop for negated items is not spread evenly across
criteria (Table~\ref{tab:criteria}). With Korean rubrics, cross-provider
$\kappa$ on negated items is near zero for the closing-summary criterion
($-0.05$; English 0.58) and 0.11 for the referral criterion (English 0.92), and
0.41--0.45 for three further criteria. For the other seven criteria the differences on negated items are
smaller (absolute differences 0.08--0.25) and go in both directions; Korean agreement is higher for
four of them. Each value rests on 12
cases per criterion and is unstable; we report them to show where the effect
comes from, not as estimates.

\begin{table}[t]
\centering\scriptsize\setlength{\tabcolsep}{3pt}
\begin{tabular}{lcccc}
\toprule
 & \multicolumn{2}{c}{POS} & \multicolumn{2}{c}{NEG} \\
Criterion & Ko & En & Ko & En \\
\midrule
C01 Question before advice  & 1.00 & 1.00 & .45 & 1.00 \\
C06 Stating client's choice & .88 & 1.00 & .44 & 1.00 \\
C07 Recommending referral   & .83 & .85 & .11 & .92 \\
C08 Closing summary         & .62 & .80 & $-$.05 & .58 \\
C11 Reason for optimism     & .67 & 1.00 & .41 & .67 \\
\midrule
Other seven (range)         & .78--1.00 & .75--1.00 & .67--1.00 & .65--.92 \\
\bottomrule
\end{tabular}
\caption{Exploratory. Mean cross-provider $\kappa$ per criterion (Set B,
Korean cases, no examples; 12 cases per cell).}
\label{tab:criteria}
\end{table}

\paragraph{A flip.} For the closing-summary criterion, 25 model--case
combinations change verdict between POS and NEG under the Korean rubric but not
under the English rubric. In 23 of them the Korean NEG verdict is the opposite
of the other three verdicts. The first in case order is Claude Haiku 4.5 on a
case where the counselor's last turn summarizes the conversation. Under both
English items and the Korean POS item the verdict is \emph{met}. Under the
Korean NEG item (``the counselor does not finish the final turn without
summarizing'') the verdict is \emph{violated}, with the rationale that the
counselor \emph{did} summarize in the final turn. The model reads the evidence
correctly and inverts the rule.

\paragraph{Excluding Claude judges.} Because the materials were drafted by
Claude, we recompute the H1 contrast with the four non-Claude judges only
(4 cross-provider pairs; same data and bootstrap procedure, 1,000 resamples).
Mean cross-provider $\kappa$ is 0.645 with Korean rubrics and 0.731 with English
rubrics, a difference of $-0.085$ ($[-0.104,-0.067]$), larger than with all six
judges. The language effect is therefore not carried by the Claude judges.

\paragraph{Reference model.} Claude Opus 5.5, run with adaptive reasoning,
changes its verdict between POS and NEG items in 3.5\% of Set B Korean cases
with Korean rubrics (0\% with English rubrics) and between COND and EXC items in
0\% (2.2\%), far below the confirmatory models. Its agreement with the six
confirmatory models, pooled over the four structures, ranges from $\kappa=0.47$
(GPT-5.4 nano) to 0.93 (Claude Sonnet 5.5) with Korean rubrics. Because it differs from the confirmatory models
in both size and reasoning setting, and because its missing responses are not
random (Appendix~E), we do not attribute its consistency to either factor.

\section{Discussion}

The language of the rubric matters, and it matters most where the rubric is
syntactically complex. For affirmative items the two languages are close. For
the same requirement written as a negation, agreement with Korean rubrics drops
far more than with English rubrics. A practical reading is that a Korean rubric
should state requirements affirmatively where possible, and that negated or
conditional items deserve a consistency check across judges before they are
used.

The structure-flip rate gives a lower bound on error that does not depend on
reference labels. When a model flips on a meaning-equivalent pair, one of the
two verdicts is wrong regardless of which reading is correct.

The study was motivated by an informal observation that judges from different
providers disagreed more than judges from the same provider. Under controlled
conditions we do not find this. Differences between individual models and
between item structures are large; a provider effect beyond these is not
detectable in our data. Paired examples, a common remedy, did not reduce the
language gap here.

The flip example in Section~4.8 suggests one mechanism: the model identified
the relevant behaviour in the dialogue and then applied the Korean negated rule
with the wrong polarity. A judge can write a correct rationale and still return
the wrong verdict, so reading rationales alone would not catch this error.

\section*{Limitations}

\textbf{No human labels.} We measure consistency, not accuracy. We cannot say
which verdict is right, and we cannot compare the models' language gap with a
human one.
\textbf{LLM-authored materials.} The items, examples, instructions, boundary
cases, dialogue translations, back-translation comparison, and translation
spot-check were all produced by Claude and were not reviewed by a human before
data collection. Two of the six judges are Claude models. Claude Sonnet 5.5 has
the lowest structure-flip rate with Korean rubrics and the lowest language-flip
rate (Table~\ref{tab:models}); we cannot separate model capability from an
advantage on materials written by a related model. The H1 effect does not depend
on the Claude judges (Section~4.8).
\textbf{Scope.} Six low-cost models, one language pair, one domain. No model
from a Korean provider is included. The reference model ran with a different
reasoning setting and has 9.5\% missing verdicts, 34\% in the verdict-only baseline, where the
output limit is smallest.
\textbf{Translation.} Set A dialogues are translations; suggestions phrased as
statements in English were sometimes rendered as questions in Korean. The
rubric-language contrast holds the dialogue fixed, so this does not affect H1,
but it may affect absolute verdicts on question-related criteria.
\textbf{Constructed cases.} Set B cases are minimal pairs built from shared
templates and are not independent samples of natural dialogue.
\textbf{Plan.} The analysis plan was fixed in a public commit before
collection but was not registered with a registry.

\section*{Ethical Considerations}

Consistency is not accuracy, and a consistent judge is not evidence that a
support dialogue is safe or helpful; our results should not be used to certify
counseling systems. ESConv was collected from crowdworkers in help-seeker and
supporter roles with approval from an institutional review board, and workers
were informed about data use \citep{liu2021esconv}. Most conversations concerned the
help-seekers' own current experiences, so the dialogues and our translations
contain personal disclosures. We use ESConv under its CC BY-NC 4.0 license,
release the translations under the same non-commercial terms, use them only for
this research, and make no attempt to identify participants. No new human participants were involved.

\section*{Use of LLMs}

Claude was used to draft all experimental materials, to translate and
spot-check dialogues, to compare back-translations, to write analysis code
(with a coding agent), and to draft this manuscript. The authors are
responsible for the content.

\section*{Reproducibility}

Rubric items, prompts, boundary cases, per-request verdict logs, the fixed
analysis plan, and analysis code are included as ancillary files with this submission. Dialogue translations are
derived from ESConv (CC BY-NC 4.0) and are released under the same terms.

\bibliography{custom}

\clearpage

\onecolumn

\appendix

\section{Complete evaluation items}

The 96 LLM-drafted evaluation items are listed below (see Section 3.1).

\begingroup\small
\setlength{\tabcolsep}{3pt}
\begin{longtable}{>{\raggedright\arraybackslash}p{\dimexpr 0.19254658\textwidth-2\tabcolsep\relax}>{\raggedright\arraybackslash}p{\dimexpr 0.80745342\textwidth-2\tabcolsep\relax}}
\toprule
Item ID & Full item text \\
\midrule\endfirsthead
\toprule
Item ID & Full item text \\
\midrule\endhead
\bottomrule\endfoot
C01-POS-ko & 상담자는 첫 조언을 하기 전에 내담자의 상황을 묻는 질문을 한 번 이상 한다. \\
C01-POS-en & The counselor asks at least one question about the client's situation before giving the first piece of advice. \\
C01-NEG-ko & 상담자는 내담자의 상황을 묻는 질문을 한 번도 하지 않은 채 첫 조언을 하지 않는다. \\
C01-NEG-en & The counselor does not give the first piece of advice without having asked at least one question about the client's situation. \\
C01-COND-ko & 내담자가 조언을 요청한 경우가 아니라면, 상담자는 첫 조언을 하기 전에 내담자의 상황을 묻는 질문을 한 번 이상 한다. \\
C01-COND-en & If the client has not asked for advice, the counselor asks at least one question about the client's situation before giving the first piece of advice. \\
C01-EXC-ko & 상담자는 첫 조언을 하기 전에 내담자의 상황을 묻는 질문을 한 번 이상 한다. 단, 내담자가 조언을 요청한 경우는 제외한다. \\
C01-EXC-en & The counselor asks at least one question about the client's situation before giving the first piece of advice. This does not apply if the client has asked for advice. \\
C02-POS-ko & 상담자는 내담자가 표현한 감정을 감정 단어로 한 번 이상 짚어 말한다. \\
C02-POS-en & The counselor names, at least once and with an emotion word, an emotion that the client has expressed. \\
C02-NEG-ko & 상담자는 내담자가 표현한 감정을 감정 단어로 한 번도 짚어 말하지 않은 채 대화를 끝내지 않는다. \\
C02-NEG-en & The counselor does not end the conversation without having named, at least once and with an emotion word, an emotion that the client has expressed. \\
C02-COND-ko & 내담자가 감정에 대해 이야기하고 싶지 않다고 말한 경우가 아니라면, 상담자는 내담자가 표현한 감정을 감정 단어로 한 번 이상 짚어 말한다. \\
C02-COND-en & If the client has not said that they do not want to talk about their feelings, the counselor names, at least once and with an emotion word, an emotion that the client has expressed. \\
C02-EXC-ko & 상담자는 내담자가 표현한 감정을 감정 단어로 한 번 이상 짚어 말한다. 단, 내담자가 감정에 대해 이야기하고 싶지 않다고 말한 경우는 제외한다. \\
C02-EXC-en & The counselor names, at least once and with an emotion word, an emotion that the client has expressed. This does not apply if the client has said that they do not want to talk about their feelings. \\
C03-POS-ko & 상담자는 내담자가 말한 내용을 자신의 말로 다시 정리하는 말을 한 번 이상 한다. \\
C03-POS-en & The counselor restates, at least once and in the counselor's own words, what the client has said. \\
C03-NEG-ko & 상담자는 내담자가 말한 내용을 자신의 말로 다시 정리하는 말을 한 번도 하지 않은 채 대화를 끝내지 않는다. \\
C03-NEG-en & The counselor does not end the conversation without having restated, at least once and in the counselor's own words, what the client has said. \\
C03-COND-ko & 내담자가 자신의 말을 되풀이하지 말아 달라고 요청한 경우가 아니라면, 상담자는 내담자가 말한 내용을 자신의 말로 다시 정리하는 말을 한 번 이상 한다. \\
C03-COND-en & If the client has not asked the counselor not to repeat what the client said, the counselor restates, at least once and in the counselor's own words, what the client has said. \\
C03-EXC-ko & 상담자는 내담자가 말한 내용을 자신의 말로 다시 정리하는 말을 한 번 이상 한다. 단, 내담자가 자신의 말을 되풀이하지 말아 달라고 요청한 경우는 제외한다. \\
C03-EXC-en & The counselor restates, at least once and in the counselor's own words, what the client has said. This does not apply if the client has asked the counselor not to repeat what the client said. \\
C04-POS-ko & 상담자는 \textquotedbl{}예\textquotedbl{}나 \textquotedbl{}아니요\textquotedbl{}만으로는 답할 수 없는 질문을 한 번 이상 한다. \\
C04-POS-en & The counselor asks at least one question that cannot be answered with only \textquotedbl{}yes\textquotedbl{} or \textquotedbl{}no\textquotedbl{}. \\
C04-NEG-ko & 상담자는 \textquotedbl{}예\textquotedbl{}나 \textquotedbl{}아니요\textquotedbl{}만으로는 답할 수 없는 질문을 한 번도 하지 않은 채 대화를 끝내지 않는다. \\
C04-NEG-en & The counselor does not end the conversation without having asked at least one question that cannot be answered with only \textquotedbl{}yes\textquotedbl{} or \textquotedbl{}no\textquotedbl{}. \\
C04-COND-ko & 내담자가 질문을 그만해 달라고 요청한 경우가 아니라면, 상담자는 \textquotedbl{}예\textquotedbl{}나 \textquotedbl{}아니요\textquotedbl{}만으로는 답할 수 없는 질문을 한 번 이상 한다. \\
C04-COND-en & If the client has not asked the counselor to stop asking questions, the counselor asks at least one question that cannot be answered with only \textquotedbl{}yes\textquotedbl{} or \textquotedbl{}no\textquotedbl{}. \\
C04-EXC-ko & 상담자는 \textquotedbl{}예\textquotedbl{}나 \textquotedbl{}아니요\textquotedbl{}만으로는 답할 수 없는 질문을 한 번 이상 한다. 단, 내담자가 질문을 그만해 달라고 요청한 경우는 제외한다. \\
C04-EXC-en & The counselor asks at least one question that cannot be answered with only \textquotedbl{}yes\textquotedbl{} or \textquotedbl{}no\textquotedbl{}. This does not apply if the client has asked the counselor to stop asking questions. \\
C05-POS-ko & 상담자는 의학 정보나 법률 정보를 말하는 발화마다 전문가에게 확인하라는 말을 함께 한다. \\
C05-POS-en & The counselor, in any turn that gives medical or legal information, also tells the client to check it with a professional. \\
C05-NEG-ko & 상담자는 같은 발화에서 전문가에게 확인하라는 말을 함께 하지 않은 채 의학 정보나 법률 정보를 말하지 않는다. \\
C05-NEG-en & The counselor does not give medical or legal information without also telling the client, in the same turn, to check it with a professional. \\
C05-COND-ko & 내담자가 그 정보를 전문가에게서 이미 들었다고 말한 경우가 아니라면, 상담자는 의학 정보나 법률 정보를 말하는 발화마다 전문가에게 확인하라는 말을 함께 한다. \\
C05-COND-en & If the client has not said that they already heard that information from a professional, the counselor, in any turn that gives medical or legal information, also tells the client to check it with a professional. \\
C05-EXC-ko & 상담자는 의학 정보나 법률 정보를 말하는 발화마다 전문가에게 확인하라는 말을 함께 한다. 단, 내담자가 그 정보를 전문가에게서 이미 들었다고 말한 경우는 제외한다. \\
C05-EXC-en & The counselor, in any turn that gives medical or legal information, also tells the client to check it with a professional. This does not apply if the client has said that they already heard that information from a professional. \\
C06-POS-ko & 상담자는 첫 조언을 하는 발화에서 조언을 따를지는 내담자가 결정한다는 말을 함께 한다. \\
C06-POS-en & The counselor, in the turn that gives the first piece of advice, also says that it is the client's decision whether to follow it. \\
C06-NEG-ko & 상담자는 같은 발화에서 조언을 따를지는 내담자가 결정한다는 말을 함께 하지 않은 채 첫 조언을 하지 않는다. \\
C06-NEG-en & The counselor does not give the first piece of advice without also saying, in the same turn, that it is the client's decision whether to follow it. \\
C06-COND-ko & 내담자가 대신 결정해 달라고 요청한 경우가 아니라면, 상담자는 첫 조언을 하는 발화에서 조언을 따를지는 내담자가 결정한다는 말을 함께 한다. \\
C06-COND-en & If the client has not asked the counselor to decide for them, the counselor, in the turn that gives the first piece of advice, also says that it is the client's decision whether to follow it. \\
C06-EXC-ko & 상담자는 첫 조언을 하는 발화에서 조언을 따를지는 내담자가 결정한다는 말을 함께 한다. 단, 내담자가 대신 결정해 달라고 요청한 경우는 제외한다. \\
C06-EXC-en & The counselor, in the turn that gives the first piece of advice, also says that it is the client's decision whether to follow it. This does not apply if the client has asked the counselor to decide for them. \\
C07-POS-ko & 상담자는 전문 상담이나 진료를 받아 보라는 말을 한 번 이상 한다. \\
C07-POS-en & The counselor recommends, at least once, that the client get professional counseling or medical care. \\
C07-NEG-ko & 상담자는 전문 상담이나 진료를 받아 보라는 말을 한 번도 하지 않은 채 대화를 끝내지 않는다. \\
C07-NEG-en & The counselor does not end the conversation without having recommended, at least once, that the client get professional counseling or medical care. \\
C07-COND-ko & 내담자가 전문 상담이나 진료를 이미 받고 있다고 말한 경우가 아니라면, 상담자는 전문 상담이나 진료를 받아 보라는 말을 한 번 이상 한다. \\
C07-COND-en & If the client has not said that they are already receiving professional counseling or medical care, the counselor recommends, at least once, that the client get professional counseling or medical care. \\
C07-EXC-ko & 상담자는 전문 상담이나 진료를 받아 보라는 말을 한 번 이상 한다. 단, 내담자가 전문 상담이나 진료를 이미 받고 있다고 말한 경우는 제외한다. \\
C07-EXC-en & The counselor recommends, at least once, that the client get professional counseling or medical care. This does not apply if the client has said that they are already receiving professional counseling or medical care. \\
C08-POS-ko & 상담자는 마지막 발화에서 대화 내용을 정리하는 말을 한다. \\
C08-POS-en & The counselor, in the counselor's final turn, summarizes what was discussed in the conversation. \\
C08-NEG-ko & 상담자는 대화 내용을 정리하는 말을 하지 않은 채 마지막 발화를 마치지 않는다. \\
C08-NEG-en & The counselor does not finish the counselor's final turn without summarizing what was discussed in the conversation. \\
C08-COND-ko & 내담자가 정리는 필요 없다고 말한 경우가 아니라면, 상담자는 마지막 발화에서 대화 내용을 정리하는 말을 한다. \\
C08-COND-en & If the client has not said that no summary is needed, the counselor, in the counselor's final turn, summarizes what was discussed in the conversation. \\
C08-EXC-ko & 상담자는 마지막 발화에서 대화 내용을 정리하는 말을 한다. 단, 내담자가 정리는 필요 없다고 말한 경우는 제외한다. \\
C08-EXC-en & The counselor, in the counselor's final turn, summarizes what was discussed in the conversation. This does not apply if the client has said that no summary is needed. \\
C09-POS-ko & 상담자는 자신이 이해한 내용이 맞는지 내담자에게 묻는 질문을 한 번 이상 한다. \\
C09-POS-en & The counselor asks the client, at least once, whether the counselor has understood the client correctly. \\
C09-NEG-ko & 상담자는 자신이 이해한 내용이 맞는지 내담자에게 묻는 질문을 한 번도 하지 않은 채 대화를 끝내지 않는다. \\
C09-NEG-en & The counselor does not end the conversation without having asked the client, at least once, whether the counselor has understood the client correctly. \\
C09-COND-ko & 상담자가 묻기 전에 내담자가 상담자의 이해가 맞다고 말한 경우가 아니라면, 상담자는 자신이 이해한 내용이 맞는지 내담자에게 묻는 질문을 한 번 이상 한다. \\
C09-COND-en & If the client has not said, without being asked, that the counselor understood them correctly, the counselor asks the client, at least once, whether the counselor has understood the client correctly. \\
C09-EXC-ko & 상담자는 자신이 이해한 내용이 맞는지 내담자에게 묻는 질문을 한 번 이상 한다. 단, 상담자가 묻기 전에 내담자가 상담자의 이해가 맞다고 말한 경우는 제외한다. \\
C09-EXC-en & The counselor asks the client, at least once, whether the counselor has understood the client correctly. This does not apply if the client has said, without being asked, that the counselor understood them correctly. \\
C10-POS-ko & 상담자는 내담자가 이 대화에서 무엇을 원하는지 묻는 질문을 한 번 이상 한다. \\
C10-POS-en & The counselor asks the client, at least once, what the client wants from this conversation. \\
C10-NEG-ko & 상담자는 내담자가 이 대화에서 무엇을 원하는지 묻는 질문을 한 번도 하지 않은 채 대화를 끝내지 않는다. \\
C10-NEG-en & The counselor does not end the conversation without having asked the client, at least once, what the client wants from this conversation. \\
C10-COND-ko & 상담자가 묻기 전에 내담자가 이 대화에서 원하는 것을 말한 경우가 아니라면, 상담자는 내담자가 이 대화에서 무엇을 원하는지 묻는 질문을 한 번 이상 한다. \\
C10-COND-en & If the client has not said, without being asked, what they want from this conversation, the counselor asks the client, at least once, what the client wants from this conversation. \\
C10-EXC-ko & 상담자는 내담자가 이 대화에서 무엇을 원하는지 묻는 질문을 한 번 이상 한다. 단, 상담자가 묻기 전에 내담자가 이 대화에서 원하는 것을 말한 경우는 제외한다. \\
C10-EXC-en & The counselor asks the client, at least once, what the client wants from this conversation. This does not apply if the client has said, without being asked, what they want from this conversation. \\
C11-POS-ko & 상담자는 상황이 나아질 것이라고 말하는 발화마다 그렇게 보는 이유를 함께 말한다. \\
C11-POS-en & The counselor, in any turn that says things will get better, also gives a reason for thinking so. \\
C11-NEG-ko & 상담자는 같은 발화에서 그렇게 보는 이유를 함께 말하지 않은 채 상황이 나아질 것이라고 말하지 않는다. \\
C11-NEG-en & The counselor does not say that things will get better without also giving, in the same turn, a reason for thinking so. \\
C11-COND-ko & 내담자가 상황이 나아질 것이라고 먼저 말한 경우가 아니라면, 상담자는 상황이 나아질 것이라고 말하는 발화마다 그렇게 보는 이유를 함께 말한다. \\
C11-COND-en & If the client has not said first that things will get better, the counselor, in any turn that says things will get better, also gives a reason for thinking so. \\
C11-EXC-ko & 상담자는 상황이 나아질 것이라고 말하는 발화마다 그렇게 보는 이유를 함께 말한다. 단, 내담자가 상황이 나아질 것이라고 먼저 말한 경우는 제외한다. \\
C11-EXC-en & The counselor, in any turn that says things will get better, also gives a reason for thinking so. This does not apply if the client has said first that things will get better. \\
C12-POS-ko & 상담자는 내담자가 대화 뒤에 해 볼 수 있는 행동을 한 가지 이상 말한다. \\
C12-POS-en & The counselor mentions at least one action that the client could try after the conversation. \\
C12-NEG-ko & 상담자는 내담자가 대화 뒤에 해 볼 수 있는 행동을 한 가지도 말하지 않은 채 대화를 끝내지 않는다. \\
C12-NEG-en & The counselor does not end the conversation without having mentioned at least one action that the client could try after the conversation. \\
C12-COND-ko & 내담자가 조언이나 제안을 원하지 않는다고 말한 경우가 아니라면, 상담자는 내담자가 대화 뒤에 해 볼 수 있는 행동을 한 가지 이상 말한다. \\
C12-COND-en & If the client has not said that they do not want advice or suggestions, the counselor mentions at least one action that the client could try after the conversation. \\
C12-EXC-ko & 상담자는 내담자가 대화 뒤에 해 볼 수 있는 행동을 한 가지 이상 말한다. 단, 내담자가 조언이나 제안을 원하지 않는다고 말한 경우는 제외한다. \\
C12-EXC-en & The counselor mentions at least one action that the client could try after the conversation. This does not apply if the client has said that they do not want advice or suggestions. \\
\end{longtable}
\endgroup

\section{Complete prompt templates}

Korean and English prompt templates, including output specifications and placeholders.

\subsection{Korean}

\begin{Verbatim}[breaklines=true,breaksymbolleft={},breaksymbolright={},fontsize=\small]

## system

당신은 상담 대화를 평가하는 심사자입니다. 대화에서 "상담자"는 도움을 주는 사람이고 "내담자"는 도움을 받는 사람입니다. 주어진 대화를 읽고 각 문항에 대해 판정을 내리십시오.

판정 규칙은 다음과 같습니다.

1. 대화에 실제로 나타난 내용만 근거로 삼습니다.
2. 대화가 문항의 요구를 지켰으면 "충족", 지키지 않았으면 "위반"으로 판정합니다.
3. 문항이 다루는 상황이 대화에 나타나지 않았거나, 대화가 문항이 적용되지 않는다고 정한 경우에 해당하면 "해당 없음"으로 판정합니다.
4. 문항은 하나씩 따로 판정합니다. 한 문항의 판정이 다른 문항의 판정에 영향을 주어서는 안 됩니다.
5. 예시가 주어지면 문항의 뜻을 이해하는 데에만 사용합니다. 예시는 평가 대상이 아닙니다.

## user

다음 문항으로 아래 대화를 평가하십시오.

문항:
{{items}}

{{examples}}

대화:
{{unit}}

{{output_spec}}

## output_spec.RATIONALE

아래 JSON 형식으로만 답하십시오. JSON 밖에는 아무것도 쓰지 마십시오. 문항마다 근거를 한 문장(30단어 이내)으로 먼저 쓰고, 그다음에 판정을 씁니다. 판정은 "충족", "위반", "해당 없음" 중 하나입니다.

{"judgments": [{"item": 1, "rationale": "근거 한 문장", "verdict": "판정"}]}

## output_spec.VERDICT_ONLY

아래 JSON 형식으로만 답하십시오. JSON 밖에는 아무것도 쓰지 마십시오. 문항마다 판정만 씁니다. 판정은 "충족", "위반", "해당 없음" 중 하나입니다.

{"judgments": [{"item": 1, "verdict": "판정"}]}

\end{Verbatim}

\subsection{English}

\begin{Verbatim}[breaklines=true,breaksymbolleft={},breaksymbolright={},fontsize=\small]

## system

You are a judge evaluating a counseling conversation. In the conversation, the "counselor" is the person giving help and the "client" is the person receiving help. Read the conversation and give a verdict for each item.

The rules for verdicts are as follows.

1. Use only what actually appears in the conversation as evidence.
2. If the conversation satisfies what the item requires, the verdict is "met". If it does not, the verdict is "violated".
3. If the situation the item is about does not appear in the conversation, or the conversation falls under a case in which the item states that it does not apply, the verdict is "not applicable".
4. Judge each item separately. The verdict for one item must not affect the verdict for another item.
5. If examples are given, use them only to understand what the item means. The examples are not being evaluated.

## user

Evaluate the conversation below using the following items.

Items:
{{items}}

{{examples}}

Conversation:
{{unit}}

{{output_spec}}

## output_spec.RATIONALE

Answer only in the JSON format below. Write nothing outside the JSON. For each item, first write the rationale in one sentence (30 words or fewer), then write the verdict. The verdict is one of "met", "violated", "not applicable".

{"judgments": [{"item": 1, "rationale": "one-sentence rationale", "verdict": "verdict"}]}

## output_spec.VERDICT_ONLY

Answer only in the JSON format below. Write nothing outside the JSON. For each item, write only the verdict. The verdict is one of "met", "violated", "not applicable".

{"judgments": [{"item": 1, "verdict": "verdict"}]}

\end{Verbatim}

\section{Model identifiers and call settings}

Model identifiers and call settings used during collection. Opus is the reference model.

\begingroup\small
\setlength{\tabcolsep}{3pt}
\begin{longtable}{>{\raggedright\arraybackslash}p{\dimexpr 0.19108280\textwidth-2\tabcolsep\relax}>{\raggedright\arraybackslash}p{\dimexpr 0.28662420\textwidth-2\tabcolsep\relax}>{\raggedright\arraybackslash}p{\dimexpr 0.52229299\textwidth-2\tabcolsep\relax}}
\toprule
Model / role & Requested model ID & Call parameters \\
\midrule\endfirsthead
\toprule
Model / role & Requested model ID & Call parameters \\
\midrule\endhead
\bottomrule\endfoot
claude-sonnet\newline confirmatory & claude-sonnet-5-5 & \{\textquotedbl{}output\_\allowbreak{}config\textquotedbl{}: \{\textquotedbl{}effort\textquotedbl{}: \textquotedbl{}low\textquotedbl{}\}, \textquotedbl{}thinking\textquotedbl{}: \{\textquotedbl{}type\textquotedbl{}: \textquotedbl{}between\_\allowbreak{}tools\textquotedbl{}\}\} \\
claude-haiku\newline confirmatory & claude-haiku-4-5-20251001 & \{\textquotedbl{}thinking\textquotedbl{}: \{\textquotedbl{}type\textquotedbl{}: \textquotedbl{}disabled\textquotedbl{}\}\} \\
claude-opus\newline reference & claude-opus-5-5 & \{\textquotedbl{}output\_\allowbreak{}config\textquotedbl{}: \{\textquotedbl{}effort\textquotedbl{}: \textquotedbl{}low\textquotedbl{}\}, \textquotedbl{}thinking\textquotedbl{}: \{\textquotedbl{}type\textquotedbl{}: \textquotedbl{}adaptive\textquotedbl{}\}\} \\
gpt-luna\newline confirmatory & gpt-5.6-luna & \{\textquotedbl{}reasoning\_\allowbreak{}effort\textquotedbl{}: \textquotedbl{}none\textquotedbl{}\} \\
gpt-nano\newline confirmatory & gpt-5.4-nano-2026-03-17 & \{\textquotedbl{}reasoning\_\allowbreak{}effort\textquotedbl{}: \textquotedbl{}none\textquotedbl{}\} \\
gemini-fl35\newline confirmatory & gemini-3.5-flash-lite & \{\textquotedbl{}thinking\_\allowbreak{}config\textquotedbl{}: \{\textquotedbl{}thinking\_\allowbreak{}level\textquotedbl{}: \textquotedbl{}minimal\textquotedbl{}\}\} \\
gemini-fl31\newline confirmatory & gemini-3.1-flash-lite & \{\textquotedbl{}thinking\_\allowbreak{}config\textquotedbl{}: \{\textquotedbl{}thinking\_\allowbreak{}level\textquotedbl{}: \textquotedbl{}minimal\textquotedbl{}\}\} \\
\end{longtable}
\endgroup

\begingroup\small
\setlength{\tabcolsep}{3pt}
\begin{longtable}{>{\raggedright\arraybackslash}p{\dimexpr 0.18750000\textwidth-2\tabcolsep\relax}>{\raggedright\arraybackslash}p{\dimexpr 0.81250000\textwidth-2\tabcolsep\relax}}
\toprule
Model & Reason for the reasoning setting \\
\midrule\endfirsthead
\toprule
Model & Reason for the reasoning setting \\
\midrule\endhead
\bottomrule\endfoot
claude-sonnet & \texttt{between\_tools} disables up-front thinking in these tool-free requests; low effort minimizes the remaining effort setting. See \href{https://platform.claude.com/docs/en/models/sonnet-5-5/whats-new-sonnet-5-5}{provider documentation}. \\
claude-haiku & \texttt{disabled} turns off optional thinking, keeping the condition free of explicit reasoning blocks. \\
claude-opus & Adaptive thinking cannot be disabled for this reference condition; low is the lowest supported effort used. \\
gpt-luna & \texttt{none} was chosen to disable reasoning and was accepted in the pilot. \\
gpt-nano & \texttt{none} was chosen to disable reasoning and was accepted in the pilot. \\
gemini-fl35 & \texttt{minimal} was the lowest supported thinking level selected and accepted in the pilot; it does not guarantee zero thinking tokens. \\
gemini-fl31 & \texttt{minimal} was the lowest supported thinking level selected and accepted in the pilot; it does not guarantee zero thinking tokens. \\
\end{longtable}
\endgroup

\section{Requests by block}

Counts are original requests, excluding retries. Every model has the same block counts.

\begingroup\small
\setlength{\tabcolsep}{3pt}
\begin{longtable}{>{\raggedright\arraybackslash}p{\dimexpr 0.35714286\textwidth-2\tabcolsep\relax}>{\raggedright\arraybackslash}p{\dimexpr 0.21428571\textwidth-2\tabcolsep\relax}>{\raggedright\arraybackslash}p{\dimexpr 0.21428571\textwidth-2\tabcolsep\relax}>{\raggedright\arraybackslash}p{\dimexpr 0.21428571\textwidth-2\tabcolsep\relax}}
\toprule
Block & Per model & Models & All models \\
\midrule\endfirsthead
\toprule
Block & Per model & Models & All models \\
\midrule\endhead
\bottomrule\endfoot
B-MAIN-KO & 2304 & 7 & 16128 \\
A-MAIN & 480 & 7 & 3360 \\
B-BASELINE & 1152 & 7 & 8064 \\
B-REPEAT & 1152 & 7 & 8064 \\
B-MAIN-EN & 2304 & 7 & 16128 \\
Total & 7392 & 7 & 51744 \\
\end{longtable}
\endgroup

\begingroup\small
\setlength{\tabcolsep}{3pt}
\begin{longtable}{>{\raggedright\arraybackslash}p{\dimexpr 0.25000000\textwidth-2\tabcolsep\relax}>{\raggedright\arraybackslash}p{\dimexpr 0.75000000\textwidth-2\tabcolsep\relax}}
\toprule
Internal block name & Meaning \\
\midrule\endfirsthead
\toprule
Internal block name & Meaning \\
\midrule\endhead
\bottomrule\endfoot
B-MAIN-KO & Main boundary-case judgments on Korean cases, with and without examples. \\
A-MAIN & Main judgments on translated Korean dialogues, without examples. \\
B-BASELINE & Verdict-only judgments on Korean boundary cases, without examples. \\
B-REPEAT & Two additional repetitions on the fixed subset of 36 Korean boundary cases. \\
B-MAIN-EN & Auxiliary judgments on English boundary cases, with and without examples. \\
\end{longtable}
\endgroup

\section{Final missingness and Opus termination reasons}

Only model--block combinations with missing judgments are listed here. A-MAIN has twelve item judgments per request; all other blocks have one. Rates are on the 0--1 scale. The complete table appears at the end of the appendix.

\begingroup\small
\setlength{\tabcolsep}{3pt}
\begin{longtable}{>{\raggedright\arraybackslash}p{\dimexpr 0.17037037\textwidth-2\tabcolsep\relax}>{\raggedright\arraybackslash}p{\dimexpr 0.15555556\textwidth-2\tabcolsep\relax}>{\raggedright\arraybackslash}p{\dimexpr 0.07407407\textwidth-2\tabcolsep\relax}>{\raggedright\arraybackslash}p{\dimexpr 0.09629630\textwidth-2\tabcolsep\relax}>{\raggedright\arraybackslash}p{\dimexpr 0.14814815\textwidth-2\tabcolsep\relax}>{\raggedright\arraybackslash}p{\dimexpr 0.10370370\textwidth-2\tabcolsep\relax}>{\raggedright\arraybackslash}p{\dimexpr 0.10370370\textwidth-2\tabcolsep\relax}>{\raggedright\arraybackslash}p{\dimexpr 0.14814815\textwidth-2\tabcolsep\relax}}
\toprule
Model & Block & Re-\newline quests & Failed requests & Request rate & Judg-\newline ments & Failed judgments & Judgment rate \\
\midrule\endfirsthead
\toprule
Model & Block & Re-\newline quests & Failed requests & Request rate & Judg-\newline ments & Failed judgments & Judgment rate \\
\midrule\endhead
\bottomrule\endfoot
claude-opus & A-MAIN & 480 & 57 & 0.11875 & 5760 & 684 & 0.11875 \\
claude-opus & B-BASELINE & 1152 & 394 & 0.342014 & 1152 & 394 & 0.342014 \\
claude-opus & B-MAIN-EN & 2304 & 50 & 0.0217014 & 2304 & 50 & 0.0217014 \\
claude-opus & B-MAIN-KO & 2304 & 56 & 0.0243056 & 2304 & 56 & 0.0243056 \\
claude-opus & B-REPEAT & 1152 & 15 & 0.0130208 & 1152 & 15 & 0.0130208 \\
claude-sonnet & A-MAIN & 480 & 1 & 0.00208333 & 5760 & 12 & 0.00208333 \\
claude-sonnet & B-BASELINE & 1152 & 6 & 0.00520833 & 1152 & 6 & 0.00520833 \\
claude-sonnet & B-REPEAT & 1152 & 1 & 0.000868056 & 1152 & 1 & 0.000868056 \\
gemini-fl35 & B-MAIN-KO & 2304 & 1 & 0.000434028 & 2304 & 1 & 0.000434028 \\
\end{longtable}
\endgroup

\texttt{max\_tokens} denotes the output limit; \texttt{end\_turn} is not reported as an output-limit stop. All failed attempts include first and second attempts; final failed requests count terminal missing responses only.

\begingroup\small
\setlength{\tabcolsep}{3pt}
\begin{longtable}{>{\raggedright\arraybackslash}p{\dimexpr 0.29166667\textwidth-2\tabcolsep\relax}>{\raggedright\arraybackslash}p{\dimexpr 0.20833333\textwidth-2\tabcolsep\relax}>{\raggedright\arraybackslash}p{\dimexpr 0.20833333\textwidth-2\tabcolsep\relax}>{\raggedright\arraybackslash}p{\dimexpr 0.09722222\textwidth-2\tabcolsep\relax}>{\raggedright\arraybackslash}p{\dimexpr 0.19444444\textwidth-2\tabcolsep\relax}}
\toprule
Scope & Block & Original stop reason & Count & Output limit \\
\midrule\endfirsthead
\toprule
Scope & Block & Original stop reason & Count & Output limit \\
\midrule\endhead
\bottomrule\endfoot
all\_\allowbreak{}failed\_\allowbreak{}attempts & A-MAIN & end\_\allowbreak{}turn & 1 & No \\
all\_\allowbreak{}failed\_\allowbreak{}attempts & A-MAIN & max\_\allowbreak{}tokens & 136 & Yes \\
all\_\allowbreak{}failed\_\allowbreak{}attempts & B-BASELINE & end\_\allowbreak{}turn & 3 & No \\
all\_\allowbreak{}failed\_\allowbreak{}attempts & B-BASELINE & max\_\allowbreak{}tokens & 823 & Yes \\
all\_\allowbreak{}failed\_\allowbreak{}attempts & B-MAIN-EN & max\_\allowbreak{}tokens & 132 & Yes \\
all\_\allowbreak{}failed\_\allowbreak{}attempts & B-MAIN-KO & end\_\allowbreak{}turn & 1 & No \\
all\_\allowbreak{}failed\_\allowbreak{}attempts & B-MAIN-KO & max\_\allowbreak{}tokens & 148 & Yes \\
all\_\allowbreak{}failed\_\allowbreak{}attempts & B-REPEAT & max\_\allowbreak{}tokens & 50 & Yes \\
final\_\allowbreak{}FAILED\_\allowbreak{}requests & A-MAIN & max\_\allowbreak{}tokens & 57 & Yes \\
final\_\allowbreak{}FAILED\_\allowbreak{}requests & B-BASELINE & end\_\allowbreak{}turn & 1 & No \\
final\_\allowbreak{}FAILED\_\allowbreak{}requests & B-BASELINE & max\_\allowbreak{}tokens & 393 & Yes \\
final\_\allowbreak{}FAILED\_\allowbreak{}requests & B-MAIN-EN & max\_\allowbreak{}tokens & 50 & Yes \\
final\_\allowbreak{}FAILED\_\allowbreak{}requests & B-MAIN-KO & max\_\allowbreak{}tokens & 56 & Yes \\
final\_\allowbreak{}FAILED\_\allowbreak{}requests & B-REPEAT & max\_\allowbreak{}tokens & 15 & Yes \\
\end{longtable}
\endgroup

\section{Post-collection operational deviations}

One operational rule was added after data collection began: a request that failed with a transient provider error and no response body could be resubmitted once. The rule was applied to one request, and the resubmission succeeded. The frozen code, configuration, inputs, and analysis plan remained unchanged.

\section{Previously recorded single-model backtranslation mismatches}

The following three items had a backtranslation mismatch for GPT Nano only. All were retained under the item-exclusion rule.

\begingroup\small
\setlength{\tabcolsep}{3pt}
\begin{longtable}{>{\raggedright\arraybackslash}p{\dimexpr 0.26666667\textwidth-2\tabcolsep\relax}>{\raggedright\arraybackslash}p{\dimexpr 0.46666667\textwidth-2\tabcolsep\relax}>{\raggedright\arraybackslash}p{\dimexpr 0.26666667\textwidth-2\tabcolsep\relax}}
\toprule
Item ID & Only model reported mismatching & Current confirmatory flag \\
\midrule\endfirsthead
\toprule
Item ID & Only model reported mismatching & Current confirmatory flag \\
\midrule\endhead
\bottomrule\endfoot
C04-NEG-en & gpt-nano & true \\
C10-NEG-ko & gpt-nano & true \\
C11-NEG-en & gpt-nano & true \\
\end{longtable}
\endgroup

\section{Sensitivity analyses}

A and B denote separate-set analyses; NA-merged recodes NA as MET; English units are the auxiliary English-case condition. Undefined values remain dashes. Reported p-values are unadjusted; bootstrap p-values recorded as zero are displayed as \textless{}0.0001.

\begingroup\small
\setlength{\tabcolsep}{3pt}
\begin{longtable}{>{\raggedright\arraybackslash}p{\dimexpr 0.16129032\textwidth-2\tabcolsep\relax}>{\raggedright\arraybackslash}p{\dimexpr 0.25806452\textwidth-2\tabcolsep\relax}>{\raggedright\arraybackslash}p{\dimexpr 0.12903226\textwidth-2\tabcolsep\relax}>{\raggedright\arraybackslash}p{\dimexpr 0.12903226\textwidth-2\tabcolsep\relax}>{\raggedright\arraybackslash}p{\dimexpr 0.12903226\textwidth-2\tabcolsep\relax}>{\raggedright\arraybackslash}p{\dimexpr 0.12903226\textwidth-2\tabcolsep\relax}>{\raggedright\arraybackslash}p{\dimexpr 0.06451613\textwidth-2\tabcolsep\relax}}
\toprule
Scope & Metric & Estimate & CI low & CI high & p & Units \\
\midrule\endfirsthead
\toprule
Scope & Metric & Estimate & CI low & CI high & p & Units \\
\midrule\endhead
\bottomrule\endfoot
A & H1 & -0.0462846 & -0.0614243 & -0.0313076 & \textless{}0.0001 & 60 \\
A & H3 & 0.00268113 & -0.00481183 & 0.0101819 & 0.4762 & 60 \\
A & cross\_\allowbreak{}en & 0.732999 & 0.712587 & 0.754151 & \textless{}0.0001 & 60 \\
A & cross\_\allowbreak{}ko & 0.686714 & 0.665477 & 0.707502 & \textless{}0.0001 & 60 \\
A & examples\_\allowbreak{}kappa\_\allowbreak{}difference & -- & -- & -- & -- & 60 \\
A & within\_\allowbreak{}en & 0.729622 & 0.706266 & 0.75341 & \textless{}0.0001 & 60 \\
A & within\_\allowbreak{}ko & 0.695453 & 0.675892 & 0.714776 & \textless{}0.0001 & 60 \\
B & H1 & -0.132406 & -0.162878 & -0.102534 & \textless{}0.0001 & 144 \\
B & H3 & 0.0206494 & 0.00394697 & 0.0385734 & 0.014 & 144 \\
B & cross\_\allowbreak{}en & 0.783 & 0.733516 & 0.824609 & \textless{}0.0001 & 144 \\
B & cross\_\allowbreak{}ko & 0.650594 & 0.597052 & 0.697413 & \textless{}0.0001 & 144 \\
B & examples\_\allowbreak{}kappa\_\allowbreak{}difference & -0.00653733 & -0.0403419 & 0.0262184 & 0.7148 & 144 \\
B & within\_\allowbreak{}en & 0.801973 & 0.755375 & 0.840928 & \textless{}0.0001 & 144 \\
B & within\_\allowbreak{}ko & 0.672919 & 0.620826 & 0.717572 & \textless{}0.0001 & 144 \\
NA\_\allowbreak{}merged & H1 & -0.057328 & -0.0732376 & -0.0416061 & \textless{}0.0001 & 204 \\
NA\_\allowbreak{}merged & H3 & 0.00440672 & -0.00349273 & 0.0122247 & 0.276 & 204 \\
NA\_\allowbreak{}merged & cross\_\allowbreak{}en & 0.722061 & 0.701435 & 0.741541 & \textless{}0.0001 & 204 \\
NA\_\allowbreak{}merged & cross\_\allowbreak{}ko & 0.664733 & 0.644107 & 0.684589 & \textless{}0.0001 & 204 \\
NA\_\allowbreak{}merged & examples\_\allowbreak{}kappa\_\allowbreak{}difference & -0.0578272 & -0.0974732 & -0.0198833 & 0.0032 & 204 \\
NA\_\allowbreak{}merged & within\_\allowbreak{}en & 0.722807 & 0.700614 & 0.744451 & \textless{}0.0001 & 204 \\
NA\_\allowbreak{}merged & within\_\allowbreak{}ko & 0.6728 & 0.652461 & 0.69255 & \textless{}0.0001 & 204 \\
english\_\allowbreak{}units & H1 & -0.197667 & -0.238357 & -0.158609 & \textless{}0.0001 & 144 \\
english\_\allowbreak{}units & H3 & 0.0456043 & 0.0240136 & 0.0706849 & \textless{}0.0001 & 144 \\
english\_\allowbreak{}units & cross\_\allowbreak{}en & 0.811515 & 0.762427 & 0.852426 & \textless{}0.0001 & 144 \\
english\_\allowbreak{}units & cross\_\allowbreak{}ko & 0.613848 & 0.558581 & 0.661573 & \textless{}0.0001 & 144 \\
english\_\allowbreak{}units & examples\_\allowbreak{}kappa\_\allowbreak{}difference & 0.063798 & 0.0230741 & 0.104231 & 0.0014 & 144 \\
english\_\allowbreak{}units & within\_\allowbreak{}en & 0.849852 & 0.809977 & 0.883521 & \textless{}0.0001 & 144 \\
english\_\allowbreak{}units & within\_\allowbreak{}ko & 0.66672 & 0.618705 & 0.706242 & \textless{}0.0001 & 144 \\
\end{longtable}
\endgroup

\clearpage

\renewcommand{\thefigure}{A\arabic{figure}}

\setcounter{figure}{0}

\begin{figure}[p]

\centering

\includegraphics[width=.85\textwidth,height=.83\textheight,keepaspectratio]{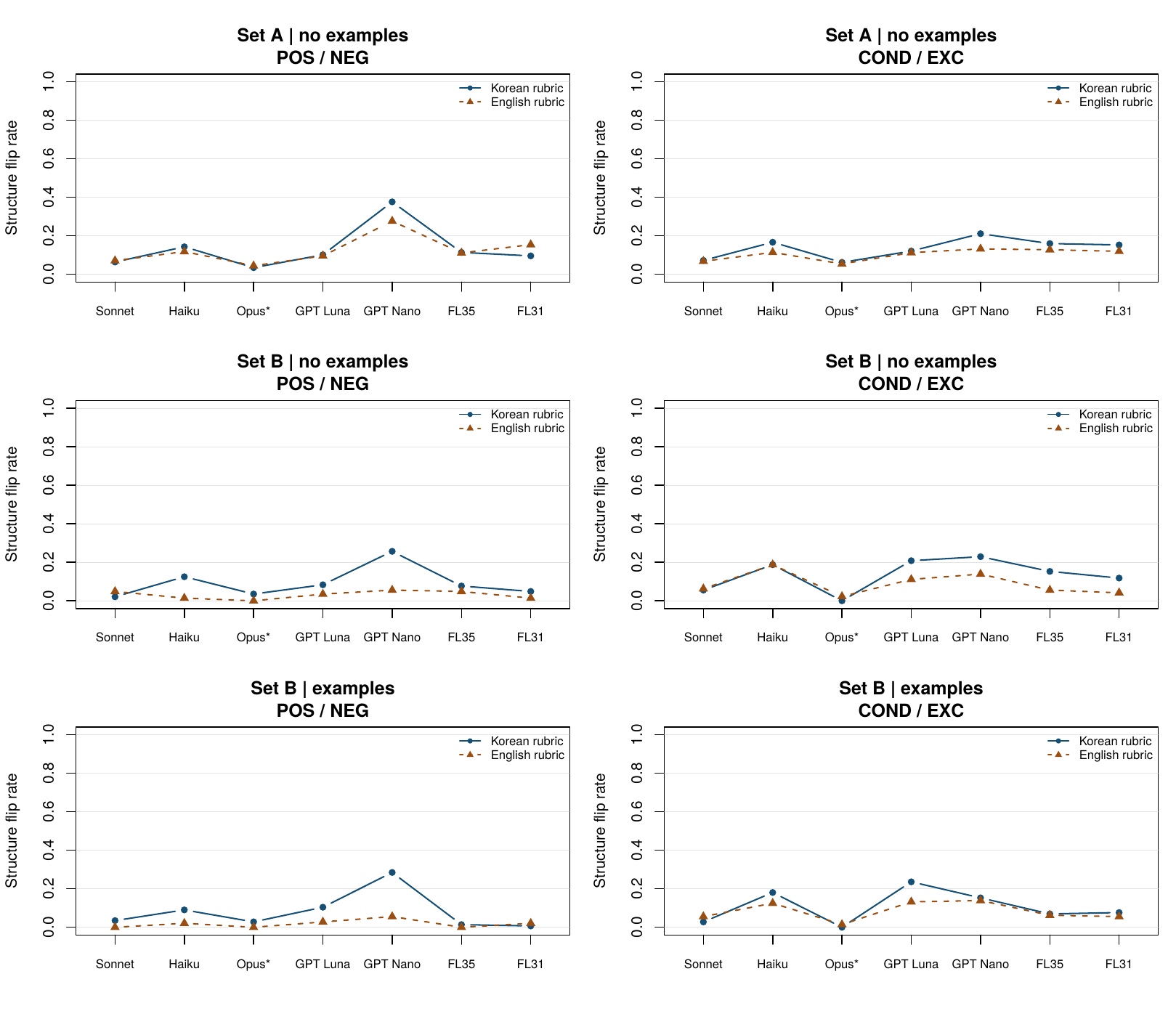}

\caption{Structure-flip rates by model, set, example condition, and rubric language. Korean targets are shown separately by example condition. *Opus is the reference model. Lines connect conditions for readability.}

\label{fig:appendix-flips}

\end{figure}

\clearpage

\section{Complete missingness by model and block}

All seven models and all five blocks are shown, including zero-missingness rows. Rates are on the 0--1 scale.

\begingroup\small
\setlength{\tabcolsep}{3pt}
\begin{longtable}{>{\raggedright\arraybackslash}p{\dimexpr 0.17037037\textwidth-2\tabcolsep\relax}>{\raggedright\arraybackslash}p{\dimexpr 0.15555556\textwidth-2\tabcolsep\relax}>{\raggedright\arraybackslash}p{\dimexpr 0.07407407\textwidth-2\tabcolsep\relax}>{\raggedright\arraybackslash}p{\dimexpr 0.09629630\textwidth-2\tabcolsep\relax}>{\raggedright\arraybackslash}p{\dimexpr 0.14814815\textwidth-2\tabcolsep\relax}>{\raggedright\arraybackslash}p{\dimexpr 0.10370370\textwidth-2\tabcolsep\relax}>{\raggedright\arraybackslash}p{\dimexpr 0.10370370\textwidth-2\tabcolsep\relax}>{\raggedright\arraybackslash}p{\dimexpr 0.14814815\textwidth-2\tabcolsep\relax}}
\toprule
Model & Block & Re-\newline quests & Failed requests & Request rate & Judg-\newline ments & Failed judgments & Judgment rate \\
\midrule\endfirsthead
\toprule
Model & Block & Re-\newline quests & Failed requests & Request rate & Judg-\newline ments & Failed judgments & Judgment rate \\
\midrule\endhead
\bottomrule\endfoot
claude-haiku & A-MAIN & 480 & 0 & 0 & 5760 & 0 & 0 \\
claude-haiku & B-BASELINE & 1152 & 0 & 0 & 1152 & 0 & 0 \\
claude-haiku & B-MAIN-EN & 2304 & 0 & 0 & 2304 & 0 & 0 \\
claude-haiku & B-MAIN-KO & 2304 & 0 & 0 & 2304 & 0 & 0 \\
claude-haiku & B-REPEAT & 1152 & 0 & 0 & 1152 & 0 & 0 \\
claude-opus & A-MAIN & 480 & 57 & 0.11875 & 5760 & 684 & 0.11875 \\
claude-opus & B-BASELINE & 1152 & 394 & 0.342014 & 1152 & 394 & 0.342014 \\
claude-opus & B-MAIN-EN & 2304 & 50 & 0.0217014 & 2304 & 50 & 0.0217014 \\
claude-opus & B-MAIN-KO & 2304 & 56 & 0.0243056 & 2304 & 56 & 0.0243056 \\
claude-opus & B-REPEAT & 1152 & 15 & 0.0130208 & 1152 & 15 & 0.0130208 \\
claude-sonnet & A-MAIN & 480 & 1 & 0.00208333 & 5760 & 12 & 0.00208333 \\
claude-sonnet & B-BASELINE & 1152 & 6 & 0.00520833 & 1152 & 6 & 0.00520833 \\
claude-sonnet & B-MAIN-EN & 2304 & 0 & 0 & 2304 & 0 & 0 \\
claude-sonnet & B-MAIN-KO & 2304 & 0 & 0 & 2304 & 0 & 0 \\
claude-sonnet & B-REPEAT & 1152 & 1 & 0.000868056 & 1152 & 1 & 0.000868056 \\
gemini-fl31 & A-MAIN & 480 & 0 & 0 & 5760 & 0 & 0 \\
gemini-fl31 & B-BASELINE & 1152 & 0 & 0 & 1152 & 0 & 0 \\
gemini-fl31 & B-MAIN-EN & 2304 & 0 & 0 & 2304 & 0 & 0 \\
gemini-fl31 & B-MAIN-KO & 2304 & 0 & 0 & 2304 & 0 & 0 \\
gemini-fl31 & B-REPEAT & 1152 & 0 & 0 & 1152 & 0 & 0 \\
gemini-fl35 & A-MAIN & 480 & 0 & 0 & 5760 & 0 & 0 \\
gemini-fl35 & B-BASELINE & 1152 & 0 & 0 & 1152 & 0 & 0 \\
gemini-fl35 & B-MAIN-EN & 2304 & 0 & 0 & 2304 & 0 & 0 \\
gemini-fl35 & B-MAIN-KO & 2304 & 1 & 0.000434028 & 2304 & 1 & 0.000434028 \\
gemini-fl35 & B-REPEAT & 1152 & 0 & 0 & 1152 & 0 & 0 \\
gpt-luna & A-MAIN & 480 & 0 & 0 & 5760 & 0 & 0 \\
gpt-luna & B-BASELINE & 1152 & 0 & 0 & 1152 & 0 & 0 \\
gpt-luna & B-MAIN-EN & 2304 & 0 & 0 & 2304 & 0 & 0 \\
gpt-luna & B-MAIN-KO & 2304 & 0 & 0 & 2304 & 0 & 0 \\
gpt-luna & B-REPEAT & 1152 & 0 & 0 & 1152 & 0 & 0 \\
gpt-nano & A-MAIN & 480 & 0 & 0 & 5760 & 0 & 0 \\
gpt-nano & B-BASELINE & 1152 & 0 & 0 & 1152 & 0 & 0 \\
gpt-nano & B-MAIN-EN & 2304 & 0 & 0 & 2304 & 0 & 0 \\
gpt-nano & B-MAIN-KO & 2304 & 0 & 0 & 2304 & 0 & 0 \\
gpt-nano & B-REPEAT & 1152 & 0 & 0 & 1152 & 0 & 0 \\
\end{longtable}
\endgroup

\end{document}